\documentclass[preprint,12pt]{elsarticle}

\usepackage{amssymb}
\usepackage{amsmath}
\usepackage{times}
\usepackage{latexsym}
\usepackage{booktabs}
\usepackage{bbding}
\usepackage{amsfonts}
\usepackage{url}
\usepackage{hyperref}
\usepackage{multirow}
\usepackage{wrapfig}
\usepackage{subcaption}

\usepackage{tikz}
\usetikzlibrary{arrows.meta, positioning}

\usepackage{lineno}

\journal{Computer Speech \& Language}
\newcommand{\method}{MiSTER-E}

\newcommand{\methodexpbold}{\textbf{Mi}xture of \textbf{S}peech-\textbf{T}ext \textbf{E}xperts for \textbf{R}ecognition of \textbf{E}motions}
\begin{document}

\begin{frontmatter}



\title{A Mixture-of-Experts Model for Multimodal Emotion Recognition in Conversations}


\author[leap]{Soumya Dutta} 
\author[leap,micro]{Smruthi Balaji}
\author[leap]{Sriram Ganapathy}
\affiliation[leap]{organization={LEAP Lab, Department of Electrical Engineering},
            addressline={IISc Bangalore}, 
            city={Bangalore},
            postcode={560012}, 
            state={Karnataka},
            country={India}}
\affiliation[micro]{organization={Microsoft},
            city={Bangalore},
            state={Karnataka},
            country={India}}

\begin{abstract}
Emotion Recognition in Conversations (ERC) presents unique challenges, requiring models to capture the temporal flow of multi-turn dialogues and to effectively integrate cues from multiple modalities. We propose \methodexpbold{} (\method{}), a modular Mixture-of-Experts (MoE) framework designed to decouple two core challenges in ERC: modality-specific context modeling and multimodal information fusion. \method{} leverages large language models (LLMs) fine-tuned for both speech and text to provide rich utterance-level embeddings, which are then enhanced through a convolutional-recurrent context modeling layer. The system integrates predictions from three experts—speech-only, text-only, and cross-modal—using a learned gating mechanism that dynamically weighs their outputs. To further encourage consistency and alignment across modalities, we introduce a supervised contrastive loss between paired speech-text representations and a KL-divergence-based regularization across expert predictions.
Importantly, \method{} does not rely on speaker identity at any stage. Experiments on three benchmark datasets—IEMOCAP, MELD, and MOSI—show that our proposal achieves $70.9\%$, $69.5\%$, and $87.9\%$ weighted F1-scores respectively, outperforming several baseline speech-text ERC systems. We also provide various ablations to highlight the contributions made in the proposed approach.
\end{abstract}






\end{frontmatter}



\section{Introduction}\label{sec:intro}
Emotion Recognition in Conversation (ERC) is the task of inferring the emotional states of speakers engaged in multi-turn interactions, often involving multiple modalities. As a key component in developing emotionally intelligent and socially aware AI systems, ERC plays a vital role in a wide array of real-world applications—ranging from empathetic dialogue agents and customer service bots to social media monitoring and mental health assessment systems~\citep{pantic2005affective,gaind2019emotion,ghosh2019emokey}. In conversational settings, emotions are not expressed in isolation; they unfold over time and are communicated through a rich combination of signals, including textual content, vocal prosody, and visual cues. These signals often interact in subtle, context-dependent ways, making it essential to capture both intra-modal dynamics and cross-modal dependencies. The inherently sequential and multimodal nature of conversations, along with the subjective and context-sensitive nature of emotions, makes ERC a complex and challenging task in affective computing.

A wide range of approaches have been proposed to advance ERC, with notable progress in areas such as contextual modeling of dialogue~\citep{hazarika2018conversational,majumder2019dialoguernn}, speaker-aware representations that capture interpersonal dependencies~\citep{hu2021mmgcn,shen2025towards}, and various strategies for multimodal fusion—from simple early concatenation~\citep{han2021improving} to more sophisticated attention-based~\citep{dutta2025llm} and tensor-based methods~\citep{zadeh2017tensor}. However, despite this progress, most existing systems adopt monolithic architectures that conflate two fundamentally distinct modeling challenges: (i) capturing the temporal context across multi-turn dialogues, and (ii) performing cross-modal fusion of information from different modalities. This entanglement can limit model performance, especially when training data is scarce—a common scenario in ERC tasks. As a result, such designs often risk overfitting to dataset-specific patterns, rather than learning generalizable emotional cues. This leads us to a central research question:
\textit{Can architectural modularity—specifically, the decoupling of context modeling from modality fusion—lead to more effective and generalizable emotion recognition in conversations?}


\textcolor{black}{In our earlier study (Dutta et al.~\citep{dutta2023hcam}),   a proposal towards addressing this question  was attempted by developing a hierarchical modeling framework for ERC. 
This work separates the modeling pipeline into two stages: first, intra-modal context modeling, followed by inter-modal fusion. 
This structure introduces a degree of architectural decoupling and was shown to improve performance.
However, a key limitation remains: in cases where there is a substantial performance gap between modalities, the overall system does not add additional value over the best uni-modal system.
}

\textcolor{black}{In this work, we introduce \methodexpbold{} (\method{}), a modular architecture that enforces a combination of experts at the logit (decision) level rather than at the feature level.    \method{} adopts a Mixture-of-Experts (MoE) formulation with three independently optimized  branches: a speech-only contextual expert, a text-only contextual expert, and a multimodal expert. The uni-modal systems capture modality-specific conversational dynamics using temporal inception networks followed by  a recurrent layer, while the multimodal expert fuses speech and text features via cross-attention and self-attention layers. The outputs from these experts are combined using a  gating mechanism, which computes a weighted sum before the softmax normalization. This logit-level fusion enables dynamic expert selection, allowing the model to  combine complementary multimodal cues.
    To further stabilize expert specialization, we incorporate contrastive and KL-based consistency losses, while supervising each expert with focal loss to address class imbalance.
    In particular, when either modality (speech or text) has a dominant advantage over the other modality (text or speech), our modality decoupling followed by MoE based combination is shown to have a unique fusion advantage compared to other prior works. 
 The following are the contributions from the work:}

\begin{itemize}

    \item \textcolor{black}{We propose \method{}, a modular framework for emotion recognition in conversations that separates modality-specific contextual modeling from multimodal integration using a logit-level Mixture-of-Experts architecture.
    }
    
    \item \textcolor{black}{We show that adaptive, per-utterance expert weighting enables effective handling of modality imbalance, allowing unimodal experts to dominate when cross-modal cues are unreliable.
    }
    
    \item\textcolor{black}{ We incorporate auxiliary training objectives, including a speech-text contrastive loss and a KL-divergence-based consistency regularization, to stabilize expert specialization.
    }
    
    \item \textcolor{black}{We demonstrate that \method{} achieves state-of-the-art performance on three benchmark ERC datasets—IEMOCAP~\cite{busso2008iemocap}, MELD~\cite{poria2019meld}, and CMU-MOSI~\cite{zadeh2016mosi}—without using speaker identity.
}
\end{itemize}

\section{Related Work}\label{sec:relatedwork}
\noindent
\textbf{Text embedding extraction}: Early approaches to ERC relied on static word embeddings such as Word2Vec~\citep{mikolov2013distributed} and GloVe~\citep{pennington2014glove}, to encode utterances~\citep{poria2015deep,zadeh2017tensor,mai2019divide}. With the advent of transformer-based language models like BERT~\citep{devlin2019bert} and RoBERTa~\citep{liu2019roberta}, ERC systems began adopting contextual encoders~\citep{hazarika2020misa,chudasama2022m2fnet,hu2023supervised}, resulting in improved text features.
Recent text-only methods~\citep{lei2023instructerc,fu2024ckerc} pose ERC as a generative task, where they fine-tune LLMs in an autoregressive manner. However, the efficient use of LLMs when text is used alongside speech is unexplored. Towards this, we adapt LLMs as text encoders for the task of text emotion recognition, thereby harnessing the power of these models and enabling fusion with other modalities.\\
\textbf{Speech embedding extraction}: Speech features in ERC have traditionally relied on hand-crafted descriptors like OpenSMILE~\citep{eyben2010opensmile} and COVAREP~\citep{degottex2014covarep}, which, while effective, often fail to generalize across datasets with diverse acoustic conditions~\citep{majumder2019dialoguernn,poria2015deep}. Recent efforts have moved toward learnable frontends such as LEAF~\citep{zeghidourleaf} and self-supervised models like HuBERT~\citep{hsu2021hubert} and wav2vec~\citep{baevski2020wav2vec}, with demonstrated success~\citep{dutta2022multimodal,lian2022smin}. However, the utilization of multi-modal LLMs for ERC is relatively unexplored. Thus, we fine tune large speech language models (SLLMs) directly for emotional inference—an approach that has not been explored for multimodal ERC before.\\
\textbf{MoE in ERC}: While Mixture-of-Experts (MoE) architectures have shown promise in scaling both language~\citep{shazeer2017outrageously,lepikhingshard} and vision models~\citep{riquelme2021scaling}, their role in ERC has been limited. One recent example, MMGAT-EMO~\citep{zhang2025moe}, combines MoE with graph attention for emotion modeling. In contrast, we use MoE to structure our model around three specialized experts—speech, text, and fused modalities — explicitly targeting the separation of context modeling and cross-modal fusion.\\
\textbf{Loss functions for ERC}: Supervised contrastive learning~\citep{khosla2020supervised} was introduced for ERC by Li et al.\citep{li2022contrast} and extended in later works \citep{song2022supervised,yu2024emotion}, using emotion class prototypes. Some approaches extend this to align modalities—e.g., aligning audio and visual cues to textual anchors~\citep{hu2022unimse}.
In contrast, we adopt a multimodal supervised contrastive loss, where intra- and inter-modality representations of utterances belonging to the same emotion class are aligned. We further introduce a consistency loss to encourage agreement among experts, reinforcing modular cooperation.\\
\textbf{LLM-based Context Modeling in ERC}: Recent works have explored the use of large language models (LLMs) within speech-processing pipelines for emotion recognition. For example, LLM-based generative error correction frameworks have been proposed to refine ASR outputs and subsequently perform downstream tasks, including emotion recognition~\citep{yang2024large}. Stepachev et al.~\cite{stepachev2024context} leverage outputs from multiple ASR systems along with fixed conversational context windows as inputs to LLMs to improve post-ASR emotion recognition. Furthermore, Zhang et al.~\cite{zhang2024improving} demonstrate that careful selection of contextual utterances can enhance emotion recognition performance, even when transcripts are derived from noisy ASR systems. These approaches highlight the growing role of LLMs in contextual and post-ASR emotion recognition settings. In contrast, MiSTER-E incorporates LLMs as embedding encoders within a modular mixture-of-experts framework, where modality-specific contextual modeling and multimodal fusion are explicitly decoupled and integrated at the decision level.

\section{Proposed Method}\label{sec:method}
\begin{figure*}[t!]
    \centering
\includegraphics[width=\linewidth, trim={0cm 2cm 0.5cm 2cm},clip]{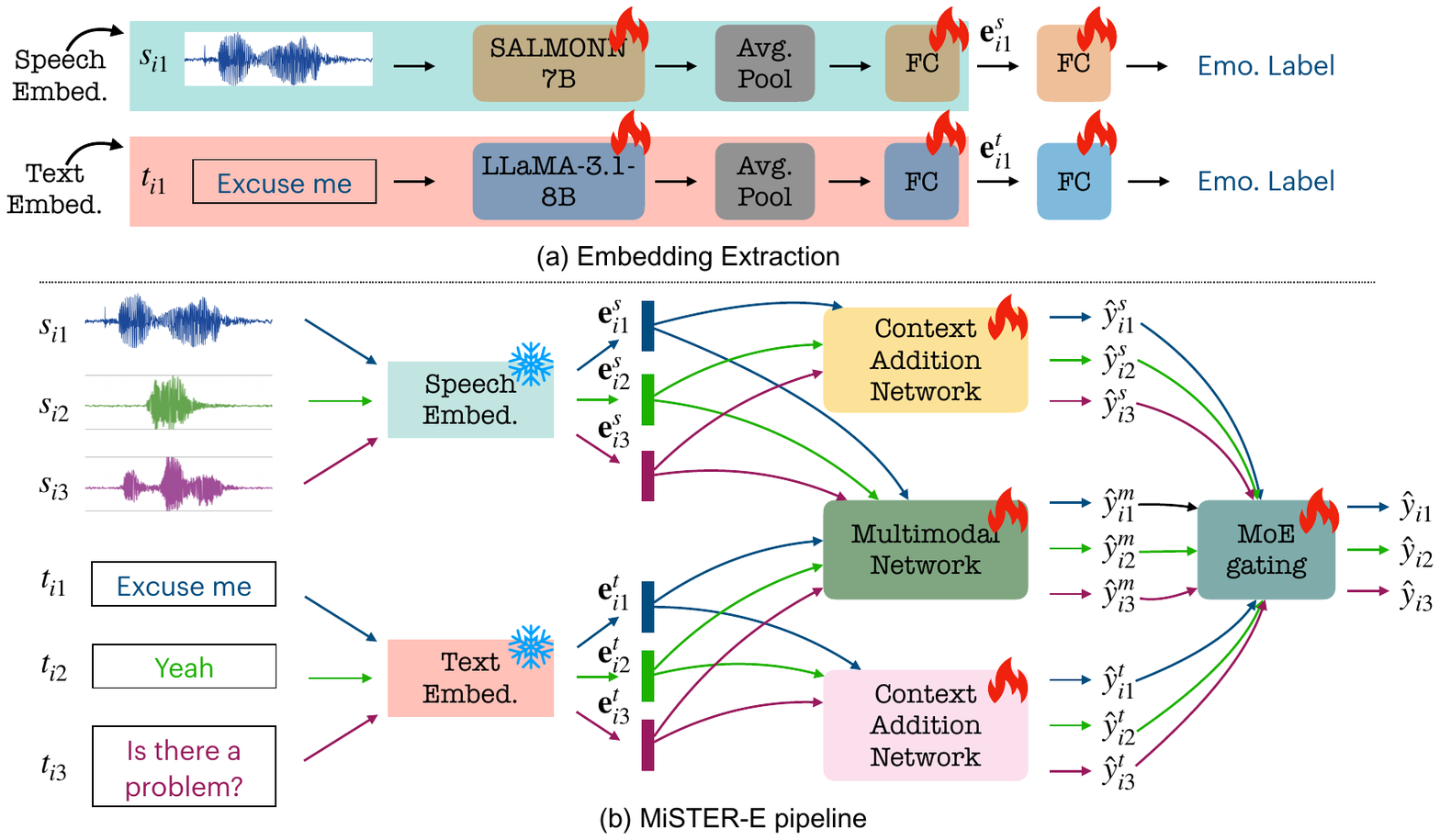}

    \caption{(a) Training of the unimodal feature extraction module (Sec.~\ref{sec:unifeat}) (b) The entire pipeline of \method{}. The speech and text embedding modules are frozen during training of the rest of the pipeline. Two context addition networks (Sec.~\ref{sec:can}) are trained for the two modalities along with a multimodal  network (Sec.~\ref{sec:fusion}). Finally, a mixture of experts gating network (Sec.~\ref{sec:moe}) is trained to predict the emotion category for each utterance.}
    \label{fig:model}

\end{figure*}

\begin{table}[t]
\centering
\resizebox{\textwidth}{!}{%
\begin{tabular}{l l}
\toprule
\textbf{\textcolor{black}{Symbol}} & \textbf{\textcolor{black}{Description}} \\
\midrule
\textcolor{black}{$\mathcal{D}$} & \textcolor{black}{Emotion Recognition in Conversations (ERC) dataset} \\
\textcolor{black}{$P$} & \textcolor{black}{Number of conversations in the dataset} \\
\textcolor{black}{$\mathcal{C}=\{\mathbf{c}_1,\mathbf{c}_2,\dots,\mathbf{c}_P\}$} & \textcolor{black}{Set of all conversations in $\mathcal{D}$} \\
\textcolor{black}{$\mathbf{c}_i$} & \textcolor{black}{The $i$-th conversation} \\
\textcolor{black}{$\mathcal{U}_i=(\mathbf{u}_{i1},\mathbf{u}_{i2},\dots,\mathbf{u}_{iN_i})$} & \textcolor{black}{Sequence of utterances in conversation $\mathbf{c}_i$} \\
\textcolor{black}{$N_i$ }& \textcolor{black}{Number of utterances in conversation $\mathbf{c}_i$} \\
\textcolor{black}{$\mathbf{u}_{ik}$} & \textcolor{black}{The $k$-th utterance in conversation $\mathbf{c}_i$} \\
\textcolor{black}{$\mathbf{s}_{ik}$ }& \textcolor{black}{Speech modality of utterance $\mathbf{u}_{ik}$} \\
\textcolor{black}{$\mathbf{t}_{ik}$ }& \textcolor{black}{Text modality of utterance $\mathbf{u}_{ik}$} \\ 
\textcolor{black}{$y_{ik}$ }& \textcolor{black}{Emotion label corresponding to utterance $\mathbf{u}_{ik}$} \\
\textcolor{black}{$\mathcal{Y}$} & \textcolor{black}{Set of emotion categories in the dataset} \\
\textcolor{black}{$\mathbf{y}_i=(y_{i1},y_{i2},\dots,y_{iN_i})$ }& \textcolor{black}{Sequence of emotion labels for conversation $\mathbf{c}_i$} \\
\midrule
\textcolor{black}{$\mathbf{e}_{ik}^s,\mathbf{e}_{ik}^t$} & \textcolor{black}{Speech and text embeddings for utterance $\mathbf{u}_{ik}$} \\
\textcolor{black}{$\hat{\mathbf{y}}_{ik}^s,\hat{\mathbf{y}}_{ik}^t,\hat{\mathbf{y}}_{ik}^m$ }& \textcolor{black}{Pre-softmax logits from speech, text, and multimodal experts} \\
\textcolor{black}{$\hat{\mathbf{y}}_{ik}$ }& \textcolor{black}{Final fused logit after mixture-of-experts gating} \\
\textcolor{black}{$\boldsymbol{\beta}_{ik}$ }& \textcolor{black}{Mixture-of-experts  weights for utterance $\mathbf{u}_{ik}$ }\\
\textcolor{black}{$\hat{\mathbf{p}}_{ik}^s,\hat{\mathbf{p}}_{ik}^t,\hat{\mathbf{p}}_{ik}^m$} & \textcolor{black}{Class probability distributions for speech, text and multimodal experts }\\
\midrule
\textcolor{black}{$\mathcal{L}_{\text{CAN}}^i$ }& \textcolor{black}{Focal loss for unimodal context addition networks} \\
\textcolor{black}{$\mathcal{L}_{\text{multi}}^i$ }& \textcolor{black}{Multimodal classification and contrastive loss} \\
\textcolor{black}{$\mathcal{L}_{\text{moe}}^i$ }& \textcolor{black}{MoE gating loss with KL-divergence consistency regularization }\\
\textcolor{black}{$\mathcal{L}_{\text{tot}}$ }& \textcolor{black}{Overall training objective} \\
\bottomrule
\end{tabular}}
\caption{\textcolor{black}{Notations used in the problem formulation and model description.}}
\label{tab:notation}
\end{table}

A block diagram of our proposed method is shown in Fig.~\ref{fig:model}.
\subsection{Problem Description} 
\textcolor{black}{Let us consider an ERC dataset $\mathcal{D}$ consisting of $P$ conversations,  $\mathcal{C}=\{\textbf{c}_1,\textbf{c}_2,\allowbreak \textbf{c}_3, \dots,\textbf{c}_P\}$, 
where each conversation $\textbf{c}_i$ consists of a set of utterances, $\mathcal{\textbf{U}}_i = \{\textbf{u}_{i1},\textbf{u}_{i2},, \dots,\textbf{u}_{iN_i}\}$, where $N_i$ refers to the number of utterances in the conversation $\textbf{c}_i$. 
In this work, only the speech and text modalities are considered, which is notated as $\textbf{u}_{ik}=\{\textbf{s}_{ik}, \textbf{t}_{ik}\}$, $k=1..N_i$,  
where $\textbf{s}_{ik}$ and $\textbf{t}_{ik}$ are speech and text data respectively.
Each utterance $\textbf{u}_{ik}$ is associated with a corresponding emotion label $y_{ik} \in \mathcal{Y}$, with $\mathcal{Y}$ denoting the label set of emotion categories in $\mathcal{D}$. 
The task of ERC is to map a sequence of utterances $\{\textbf{u}_{i1}, \textbf{u}_{i2},\dots,\textbf{u}_{iN_i}\}$ to their corresponding labels $\mathbf{y}_i=\{y_{i1},y_{i2},\dots,y_{iN_i}\}$. A summary of these notations is shown in Table~\ref{tab:notation}.}

\subsection{Unimodal Feature Extraction}\label{sec:unifeat}
\textbf{Text embeddings}: While large language models (LLMs) excel in text generation, their use in emotion recognition has been limited. Prior work typically uses them in autoregressive mode or as frozen feature extractors~\citep{behnamghaderllm2vec}.\\
In this work, we fine-tune an LLM—specifically, \texttt{LLaMA-3.1-8B}—to act as a text encoder rather than a generator. Each utterance transcript $\mathbf{t}_{ik}$ is tokenized and processed through the LLM. Token-level hidden states are then pooled and passed through a two-layer feedforward classifier trained with task-specific supervision. To preserve the pretraining knowledge while enabling efficient adaptation, we apply LoRA~\citep{hu2022lora} to fine-tune the weights.
We denote the resulting text embedding as:
\begin{equation}
    \mathbf{e}_{ik}^{t} = \texttt{Text-Embed}(\textbf{t}_{ik})
\end{equation}
where $\mathbf{e}_{ik}^t$ is extracted from the first fully connected layer of the classifier.
\\
\textbf{Speech embeddings}: For speech, we adopt a similar approach using \texttt{SALMONN\allowbreak-7B}~\citep{tangsalmonn}, a speech large language model (SLLM) comprising a speech encoder, Q-former~\citep{li2023blip}, and an LLM backbone.
For ERC, we fine-tune this model by updating the Q-former, the LLM, and a classification head with LoRA. This allows the system to learn emotionally salient acoustic patterns while retaining the semantic features of the LLM backbone. Given a speech signal $\mathbf{s}_{ik}$, we extract its representation as:
\begin{equation}
    \mathbf{e}_{ik}^{s} = \texttt{Speech-Embed}(\textbf{s}_{ik})
\end{equation}
where the embedding is taken from the first fully connected layer of the speech classification head.
Comparisons with alternative LLM and SLLM variants are provided in Sec.~\ref{sec:others}.

\subsection{Conversational Modeling}\label{sec:coer}
For a conversation $\textbf{c}_i$, the text and speech embedding sequences (Sec.~\ref{sec:unifeat}) are denoted by $\mathbf{E^t_i}=(\mathbf{e}_{i1}^{t},\dots,\mathbf{e}_{iN_i}^{t})$ and $\mathbf{E^s_i}=(\mathbf{e}_{i1}^{s},\dots,\mathbf{e}_{iN_i}^{s})$ respectively.
\begin{figure*}[t!]
    \centering
\includegraphics[width=0.75\textwidth,trim={2cm 2.3cm 1cm 3cm},clip]{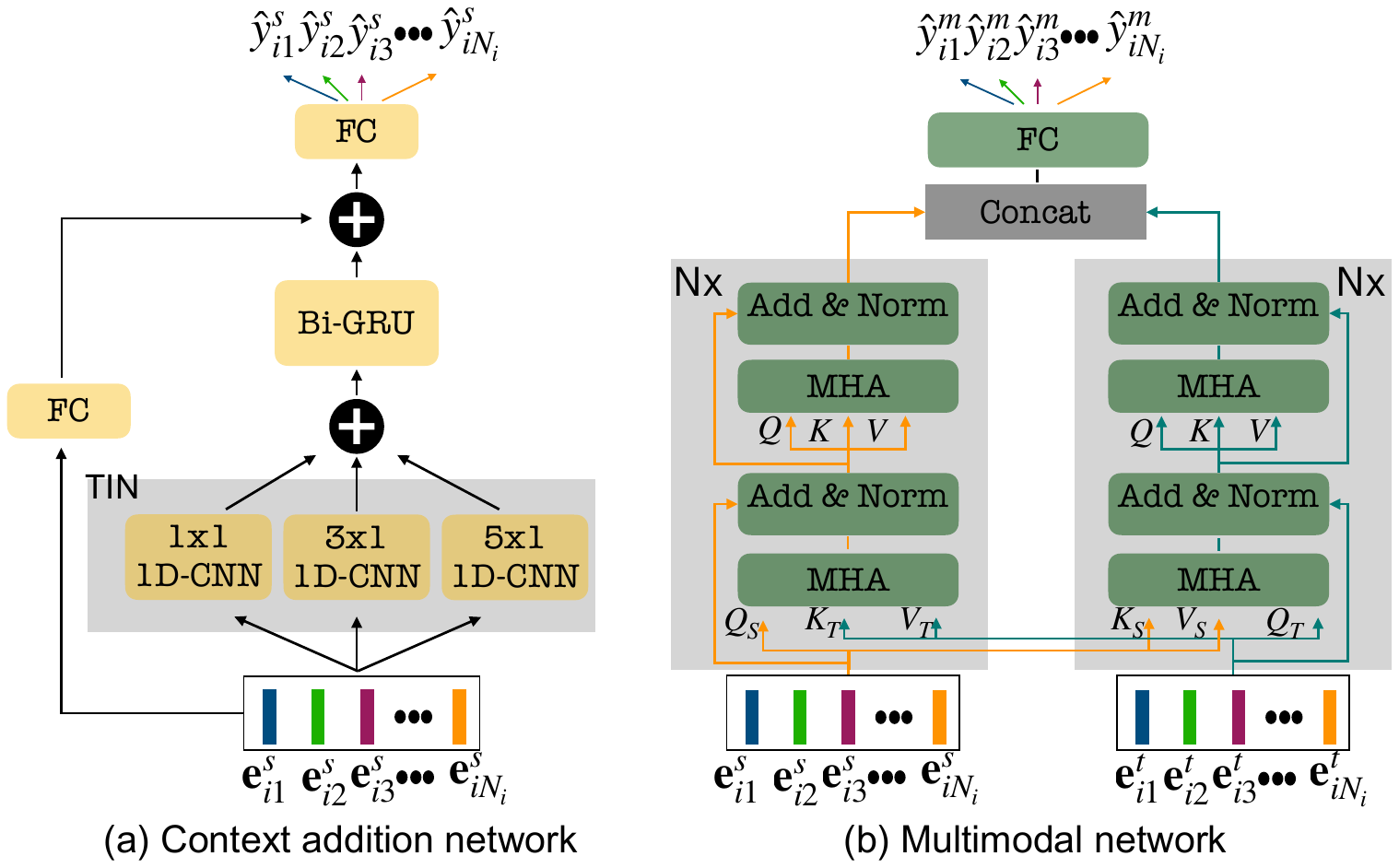}

    \caption{(a) The context addition network (for the speech modality) and (b) the multimodal network used in \method{}. The inputs to both the blocks are derived from the uni-modal feature extractor modules. \texttt{TIN} stands for Temporal Inception Network, \texttt{MHA} stands for multi-head attention.}
    \label{fig:subparts}
\end{figure*}
As shown in Figure~\ref{fig:model}, the fine-tuned speech and text embedding extractors are frozen.

\subsubsection{Context Addition Network}\label{sec:can}
To make utterance representations context-aware, we introduce a Context Addition Network (\texttt{CAN}) that enhances both text and speech embeddings (Sec.~\ref{sec:unifeat}) with conversational context. At its core is a Temporal Inception Network (\texttt{TIN}), which applies 1D convolutions with kernel sizes of $1$, $3$, and $5$ to capture short-range dependencies—simulating varying receptive fields across local utterance neighborhoods. \textcolor{black}{Inspired by the Inception architecture~\citep{szegedy2015going} originally developed for image classification, TIN enables multi-scale temporal context modeling within conversations.}\\
However, emotional signals in dialogue often evolve over longer durations. To capture such global dependencies, we append a Bi-GRU layer, allowing the model to integrate information across the entire conversational span.
A residual connection links the original embedding with its context-enhanced version, enabling additive refinement while preserving the semantic grounding of the base LLM features.
 Finally, a FC layer is used to map each utterance $\textbf{u}_{ik}$ to its corresponding emotion category $y_{ik}$.
 We train two similar networks for the speech and text modalities, respectively. These operations are denoted as:
\begin{eqnarray}\label{eq:can_s}
    \mathbf{\hat{y}^s_i} = (\hat{y}_{i1}^s, \dots,\hat{y}_{iN_i}^s)=\texttt{CAN}(\mathbf{E_i^s})\\ 
    \mathbf{\hat{y}^t_i} =(\hat{y}_{i1}^t, \dots,\hat{y}_{iN_i}^t)=\texttt{CAN}(\mathbf{E_i^t})
\end{eqnarray}
A schematic of the speech-side \texttt{CAN} is shown in Fig.~\ref{fig:subparts}(a).

\subsubsection{Multimodal Network}\label{sec:fusion}
To integrate information across modalities, we design a fusion module based on cross-attention between speech and text. This mechanism allows the model to align semantic cues from text with affective signals from audio.
The speech and text embeddings, $\mathbf{E_i^s}$ and $\mathbf{E_i^t}$, are first projected into query, key, and value spaces. Cross-attention is then applied bidirectionally:
\begin{eqnarray}
    \mathbf{M^{t\rightarrow s}_i} = \texttt{LN}(\texttt{FC}(\mathbf{E_i^s})+\texttt{MHA}(\mathbf{Q_i^s},\mathbf{K_i^t},\mathbf{V_i^t})) \\
    \mathbf{M^{s\rightarrow t}_i} = \texttt{LN}(\texttt{FC}(\mathbf{E_i^t})+\texttt{MHA}(\mathbf{Q_i^t},\mathbf{K_i^s},\mathbf{V_i^s}))
\end{eqnarray}
Here, \texttt{MHA} and \texttt{LN} refer to multi-head attention and layer normalization, respectively.\\
While this enables inter-modal alignment, it overlooks cues from the conversational context that are essential for ERC. To address this, we apply modality-specific self-attention layers over the aligned representations, capturing temporal context across the conversation: 
\begin{eqnarray}\label{eq:multspeech}
    \mathbf{M^{s}_i} = \texttt{Self-attn.}( \mathbf{M^{t\rightarrow s}_i}) \\ 
 \mathbf{M^{t}_i} = \texttt{Self-attn.}(\mathbf{M^{s\rightarrow t}_i})
\end{eqnarray}
The outputs $\mathbf{M^{s}_i}$ and $\mathbf{M^{t}_i}$ are concatenated and passed through a fully connected (FC) layer:
\begin{equation}\label{eq:mult}
    \mathbf{\hat{y}^m_i} =(\hat{y}^m_{i1},\hat{y}^m_{i2}, \dots,\hat{y}^m_{iN_i})=\texttt{FC}([\mathbf{M^{s}_i};\mathbf{M^{t}_i}])
\end{equation}
An overview is presented in Fig.~\ref{fig:subparts}(b).

\subsection{Mixture-of-Experts Gating}\label{sec:moe}
\textcolor{black}{We obtain pre-softmax logits from the three experts for each utterance $u_{ik}$:
the speech expert ($\mathbf{\hat{y}^s_{ik}}$), the text expert ($\mathbf{\hat{y}^t_{ik}}$),
and the multimodal expert ($\mathbf{\hat{y}^m_{ik}}$).} \\
 \textcolor{black}{To combine the logits, we dynamically fuse the experts' logits using a gating mechanism that learns to weigh them.  To compute these weights, we concatenate the logits and feed them into a fully connected (FC) layer:}
\begin{equation}\label{eq:moegate}
\textcolor{black}{\mathbf{g}_{ik} = \texttt{FC}([\mathbf{\hat{y}^s_{ik}}; \mathbf{\hat{y}^t_{ik}}; \mathbf{\hat{y}^m_{ik}}])}
\end{equation}
\textcolor{black}{The gating network outputs $\mathbf{g}_{ik} \in \mathbb{R}^{3}$, which is transformed via a softmax operation to produce adaptive mixture weights ${\boldsymbol \beta}_{ik}$.}

\textcolor{black}{The final logit is computed as a weighted sum of the expert predictions:}

\begin{equation}\label{eq:final_pred}
\textcolor{black}{\mathbf{\hat{y}_{ik}} = \beta_{ik}^s \cdot \mathbf{\hat{y}^s_{ik}} + \beta_{ik}^t \cdot \mathbf{\hat{y}^t_{ik}} + \beta_{ik}^m \cdot \mathbf{\hat{y}^m_{ik}}}
\end{equation} 
 
 \textcolor{black}{A softmax operation is applied to $\mathbf{\hat{y}_{ik}}$ to obtain the final predictive distribution for utterance $u_{ik}$.
This gating network is trained end-to-end, empowering the model to flexibly integrate modalities by emphasizing the most reliable expert.}

\subsection{Model Training}
\subsubsection{Loss Function}
ERC is typically characterized by severe class imbalance, where  rare emotion classes are often misclassified~\citep{poria2019emotion}. To address this, we employ the focal loss~\citep{lin2017focal}, which modulates the contribution of each training example based on its complexity,  reducing the relative loss for well-classified examples while focusing on hard (possibly minority class) instances.
In \method{}, this loss is applied during the training of the uni-modal embedding extractors  for text and speech as well as their respective context addition networks (CANs). The loss for the speech and text CAN networks is given by:
\begin{equation}
\mathcal{L}_{\text{CAN}}^i = \sum_{k=1}^{N_i} \texttt{FL}(\hat{y}_{ik}^s, y_{ik}) + \sum_{k=1}^{N_i} \texttt{FL}(\hat{y}_{ik}^t, y_{ik})
\end{equation}
Here, $\texttt{FL}(\cdot)$ represents the focal loss function. 
Description of the focal loss and related ablations are provided in Sec.~\ref{sec:focal}.

\subsubsection{Multimodal Contrastive Loss}
In many cases, speech and text provide complementary signals about a speaker’s emotion. To exploit this, we incorporate a supervised contrastive loss that structures the joint representation space based on emotion labels, drawing together utterances with shared emotional intent across modalities.  
Consider the multimodal speech and text representations denoted by $\mathbf{M_i^s}=\{\mathbf{m_{i1}^s}, \mathbf{m_{i2}^s}, \mathbf{m_{i3}^s},\dots,\mathbf{m_{iN_i}^s}\}$ and $\mathbf{M_i^t}=\{\mathbf{m_{i1}^t}, \mathbf{m_{i2}^t}, \mathbf{m_{i3}^t},\dots,\mathbf{m_{iN_i}^t}\}$ respectively. These embeddings  are batched to get, 
\begin{equation}
    \mathbf{Z_i}=\{\mathbf{\tilde{m}_{i1}^s},\dots,\mathbf{\tilde{m}_{iN_i}^s},\mathbf{\tilde{m}_{i1}^t},\dots,\mathbf{\tilde{m}_{iN_i}^t}\}
\end{equation}
where \textcolor{black}{$\mathbf{\tilde{m}_{ik}^\cdot}=\frac{\mathbf{m_{ik}^\cdot}}{||\mathbf{m_{ik}^\cdot}||}$} denotes the normalized embeddings. 
Let $\mathbf{z}_a \in \mathbf{Z_i}$ for $a \in \{1,2,\dots,2N_i\}$, and let $y_a$ be the emotion label associated with $\mathbf{z}_a$.
The contrastive loss   is given by:
\begin{equation}\label{eq:con}
\mathcal{L}^\text{i}_{\text{con}} =
  \sum_{a=1}^{2N_i} \frac{-1}{|P(a)|}
\sum_{p \in P(a)} \log \frac{\exp\left(\mathbf{z}_a^\top \mathbf{z}_p / \tau\right)}{\sum\limits_{\substack{q=1 \\ \ q \ne a}}^{2N_i} \exp\left(\mathbf{z}_a^\top \mathbf{z}_q / \tau\right)}
\end{equation}
where $\tau$ is the temperature and $P(a)=\{p \in \{1,2,\dots,2N_i\} \backslash \{a\}|y_p=y_a\}$ is the set of positives for anchor $\mathbf{z}_a$. This objective pulls together utterances with the same emotion class across both speech and text, while pushing apart other samples, thereby guiding the model to learn emotionally coherent, modality-invariant representations.\\
\textbf{Total multimodal loss}: The final loss used to train the multimodal model combines the classification and contrastive objectives:
\begin{equation}\label{eq:tot_mult}
\mathcal{L}_{\text{multi}}^i = \sum_{k=1}^{N_i} \texttt{FL}(\hat{y}_{ik}^m, y_{ik}) + \lambda \mathcal{L}_{\text{con}}^i
\end{equation}
where $\hat{y}_{ik}^m$ denotes the multimodal prediction, and $\lambda$ controls the contribution of the contrastive loss. 
\subsubsection{MoE Gating Loss}
To train the mixture-of-experts (MoE) gating network, we combine two objectives: (i) focal loss for emotion classification, and (ii) a regularization term to promote consistency among the expert predictions. Specifically, we enforce similarity in the predicted distributions of the three experts—speech-only, text-only, and multimodal—using the Kullback-Leibler (KL) divergence.
The total loss for the MoE layer is:
\begin{equation}\label{eq:kl}
\mathcal{L}_{\text{moe}}^i = \sum_{k=1}^{N_i} \texttt{FL}(\hat{y}_{ik}, y_{ik}) + \alpha \cdot \mathcal{L}_{\text{KL}}^i
\end{equation}
where $\hat{y}_{ik}$ is defined in Eq.~\ref{eq:final_pred}, $\alpha$ controls the strength of the consistency regularization, and the KL term is given by:
\begin{equation}
\textcolor{black}{\mathcal{L}_{\text{KL}}^i = \sum_{k=1}^{N_i} \Big[ \texttt{KL}(\hat{p}_{ik}^m || \hat{p}_{ik}^s) +
\texttt{KL}(\hat{p}_{ik}^m || \hat{p}_{ik}^t) \Big]}
\end{equation}
\textcolor{black}{Here, $\hat{p}_{ik}^{\cdot}$ denotes the class probability distribution obtained by applying a softmax function to the corresponding expert logits $\hat{y}_{ik}^{\cdot}$. The KL consistency loss encourages the unimodal speech and text experts to produce predictive distributions that are aligned with the multimodal expert. Since the multimodal expert jointly models both modalities, we treat its output distribution as a stronger supervisory signal and therefore, employ a directional KL divergence in this manner.
}

\subsubsection{Total Loss}
The context addition networks, the multimodal network, and the MoE gating layer are trained together. The total loss is:
\begin{equation}\label{eq:all}
 \mathcal{L}_{\text{tot}} = \sum_{i=1}^{P} \bigg[
\mathcal{L}_{\text{CAN}}^i + \mathcal{L}_{\text{moe}}^i \\
+ \mathcal{L}_{\text{multi}}^i \bigg]
\end{equation}

\section{Experiments}\label{sec:experiments}

\subsection{Datasets}
We evaluate \method{} on \textcolor{black}{three ERC datasets - IEMOCAP~\citep{busso2008iemocap}, MELD~\citep{poria2019meld}, and CMU-MOSI~\cite{zadeh2016mosi}}. \\
\textbf{IEMOCAP} consists of conversational data split into $5$ sessions, $151$ dialogues and $7433$ utterances. Following prior work~\citep{lian2022smin}, we consider session $5$ for testing, while session $1$ is used for validation. The remaining $3$ sessions are used for training. Each utterance is classified as one of six emotions: ``angry'', ``happy'', ``sad'', ``frustrated'', ``excited'' and ``neutral''. Thus, we use $92$ conversations for training, $28$ conversations for validation and the $31$ conversations in Session $5$ for testing. There are a total of $5810$ utterances for training and validation, while $1623$ utterances are used for testing the model.\\
\textbf{MELD} is a multi-party conversational dataset consisting of $1433$ dialogues with $13708$ utterances from the TV show \textit{Friends}. This dataset has pre-defined train, validation, and test splits which are used in this work. Each utterance is categorized as one of seven emotion classes: ``angry'', ``joy'', ``sadness'', ``fear'', ``disgust'', ``surprise'' and ``neutral''. The training and validation data together amount to $1153$ conversations ($11098$ utterances), while the test data comprises $280$ conversations ($2610$ utterances).\\
\textcolor{black}{\textbf{CMU-MOSI} consists of $93$ monologues comprising $2{,}199$ utterances, each annotated with a sentiment score in the range $[-3, 3]$. Following prior work, we formulate the task as a binary classification problem, where utterances with sentiment scores in the range $[-3, 0)$ are labeled as negative and those in the range $[0, 3]$ as positive. We adopt the same dataset partitioning protocol as reported in Lian et al.~\cite{lian2022smin}: the first $62$ monologues are used for training and validation, and the remaining $31$ monologues are reserved for testing. Among the $62$ monologues, $49$ are used for training and $13$ for validation. This results in $1{,}188$ training utterances, $325$ validation utterances, and $686$ test utterances.}

\subsection{Implementation details}
The two unimodal feature extractors (\texttt{SALMONN-7B} and \texttt{LLaMA-3.1-8B}) are trained using LoRA with a rank of $8$ and the scaling parameter of $32$ with a dropout of $0.1$. Both models are trained with a batch size of $8$ and a learning rate of $1e-5$, with the focal loss. The hidden dimension in the FC (Sec.~\ref{sec:unifeat}) is set to $2048$. For the rest of the \method{} pipeline, we use a batch size of $8$ for IEMOCAP, MOSI and $32$ for MELD. The learning rate is set to $1e-5$ for all datasets. For MELD and MOSI, $\lambda$ (Eq.~\ref{eq:tot_mult}) is set to $1$ , while for IEMOCAP, it is set to $2$. The consistency regularization term,
$\alpha$ (Eq.~\ref{eq:kl}), is set to $0.1$ for IEMOCAP and MOSI, while it is kept at $1e-3$ for MELD (Sec.~\ref{sec:comb}).
We report the weighted F1-score on the test data as the performance metric with $3$ random initializations. \\
All our experiments are run using a NVIDIA RTX A6000  GPU card with Pytorch 2.7.0 \footnote{\url{https://pytorch.org/blog/pytorch-2-7/}} and CUDA 12.6. The number of trainable parameters in our model amounts to $97$M ($8$M each for the LLM/SLLM LoRA training and $81$M for the conversational modeling). Training the LLM takes about $10$ minutes per epoch, while the SLLM takes about $20$ minutes per epoch. The rest of the model, following feature encoders takes approximately $10$ minutes for $100$ epochs. We mention further hyperparameter choices for both datasets below:
\begin{itemize}
    \item The BiGRU used in the \texttt{CAN} network (Sec.~\ref{sec:can}) has a hidden dimension of $512$ and we use $3$ layers for IEMOCAP, MOSI, and $2$ for MELD. We use a dropout of $0.2$ between each fully connected layer for regularization.
    \item For the multimodal fusion network, we use $4$ layers with hidden dimension of $120$ and $4$ attention heads. We use a dropout regularization of $0.5$ in the multimodal fusion network. A fusion network with the same parameters is used for both the datasets. The temperature parameter for the contrastive loss, used for the fusion network (Eq.~\ref{eq:con}), is kept at $1$ for IEMOCAP, MOSI, and $0.05$ for MELD.
    \item While training the context addition networks, the multimodal network, and the MoE gating network, we use gradient clipping with norm $1.0$ for both the datasets and train for $100$ epochs. While training the LLM and the SLLM feature encoders, we run for $50$ epochs for both datasets.\footnote{Code for \method{} is available at \url{https://github.com/iiscleap/MiSTER-E}.}
\end{itemize}

\begin{table*}[t!]
\centering
\resizebox{0.75\textwidth}{!}{%
\begin{tabular}{@{}l|c|c|c@{}}
\toprule
\textbf{Method} 
& \textbf{IEMOCAP ($6$-class)} 
& \textbf{MELD ($7$-class)} 
& \textcolor{black}{\textbf{MOSI ($2$-class)}} \\
\midrule
SMIN~\citep{lian2022smin}$^{\#\#}$ & $\underline{70.5}\%$ & $63.7\%$ & \textcolor{black}{$79.7\%$} \\
MultiEmo~\citep{shi2023multiemo}$^{\#}$ & $66.9\%^{\pm2.0}$ & $65.3\%^{\pm0.5}$ & \textcolor{black}{$83.7\%^{\pm0.3}$} \\
HCAM~\citep{dutta2023hcam} & $\underline{70.5}\%$ & $65.8\%$ & \textcolor{black}{$\underline{85.8\%}$} \\
DF-ERC~\citep{li2023revisiting} & $69.5\%$ & $64.5\%$ & \textcolor{black}{--} \\
DER-GCN~\citep{ai2024gcn} & $64.7\%$ & $62.6\%$ & \textcolor{black}{--} \\
ELR-GNN~\citep{shou2024efficient} & $64.4\%$ & $63.2\%$ & \textcolor{black}{--} \\
Mamba-like-model~\citep{shou2024revisiting} & $70.2\%$ & $65.6\%$ & \textcolor{black}{--} \\
TelME~\citep{yun2024telme} & $69.3\%$ & $\underline{67.2\%}$ & \textcolor{black}{--} \\
CFN-ESA~\citep{li2024cfn} & $68.7\%$ & $\underline{67.2}\%$ & \textcolor{black}{--} \\
MMGAT-EMO~\citep{zhang2025moe}$^{\#}$ & $65.5\%^{\pm0.6}$ & $66.1\%^{\pm 0.4}$ & \textcolor{black}{$84.3\%^{\pm0.4}$} \\
\midrule
\method{} 
& $\mathbf{70.9\%}^{\pm0.2}$ 
& $\mathbf{69.5\%}^{\pm0.3}$ 
& \textcolor{black}{$\mathbf{87.9\%}^{\pm 0.2}$} \\
\bottomrule
\end{tabular}}
\caption{Comparison of different methods on IEMOCAP, MELD, and MOSI datasets using weighted-F1 scores. All reported results are for speech-text systems. $^{\#\#}$ We report the numbers when no external emotional data is used for training SMIN. $^{\#}$ We ran the public implementations provided by the authors under our experimental settings. Superscripted results denote the mean and standard deviation over $3$ random initializations, where applicable.}
\label{tab:method_comparison}
\end{table*}

\subsection{Comparison with prior work}
Table~\ref{tab:method_comparison} presents a comparison of the proposed approach with several baseline methods. The competing approaches span a broad spectrum of modeling paradigms, including graph-based frameworks~\cite{ai2024gcn,shou2024efficient}, a state-space formulation~\cite{shou2024revisiting}, and a mixture-of-experts (MoE) based model~\cite{zhang2025moe}. As observed, the proposed method establishes new state-of-the-art performance on all the three datasets, attaining a weighted F1 score of $\mathbf{70.9}\%$ on IEMOCAP, $\mathbf{69.5}\%$ on MELD, \textcolor{black}{and $\mathbf{87.9\%}$ on CMU-MOSI}.

It is also pertinent to highlight that, for IEMOCAP, the absence of a standardized validation set has often resulted in prior studies performing model selection directly on the test set. Such a practice can artificially inflate performance metrics and limit reproducibility. To mitigate this issue, all experiments in this work (for which official implementations were available) were conducted using our own train, validation, and test splits. Consequently, certain IEMOCAP results reported in Table~\ref{tab:method_comparison} may differ from those originally published.



\begin{figure}[t!]
    \centering
\includegraphics[width=0.45\textwidth,trim={1cm 6cm 15cm 2cm},clip]{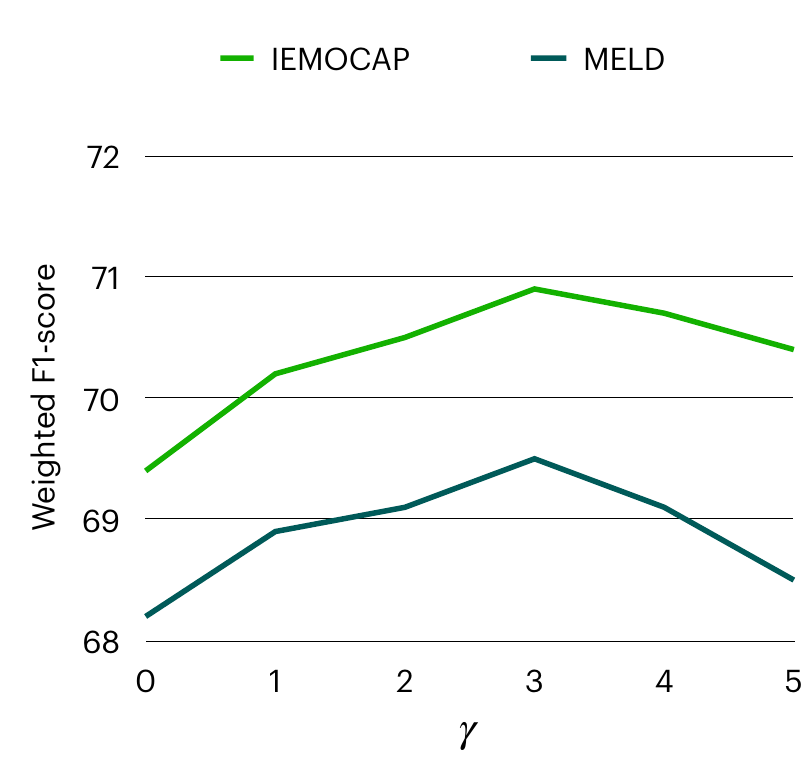}

    \caption{Performance of \method{} with different values of the focal loss hyperparameter.}
    \label{fig:focal}

\end{figure}
\subsection{Focal Loss}\label{sec:focal}
Focal loss~\cite{lin2017focal} is generally used to handle class imbalance. Since both ERC datasets considered in this paper are imbalanced, we use focal loss in place of the cross-entropy loss. Denoting the predicted probability for the sample with logits $\hat{y}_{ik}^\cdot$ as $\hat{p}_{ik}^\cdot$, the focal loss   is given as,
\begin{equation}\label{eq:focal}
\texttt{FL}(\hat{y}_{ik}^\cdot, y_{ik}) = -\sum_{c=1}^{\mathcal{Y}}y_{ik}^c(1-\hat{p}_{ik}^{\cdot c})^\gamma \log (\hat{p}_{ik}^{\cdot c})
\end{equation}
where $y_{ik}^c=1$ for the true class and $0$ otherwise.  $\hat{p}_{ik}^{\cdot c}$ is the predicted probability for class $c$ for the sample with logits $\hat{y}_{ik}^{\cdot}$ ($\cdot$ is replaced by $s$, $t$ or $m$) and $\gamma$ is the hyperparameter associated with the focal loss.

The performance of our proposed method for different values of $\gamma$ (Eq.~\ref{eq:focal}) is shown in Fig.~\ref{fig:focal}. We note that for cross-entropy loss ($\gamma=0$), the performance drops for both datasets, indicating the need for utilizing a loss function more suited for imbalanced datasets. The value of $\gamma$ is set to $3$ for both datasets.

\subsection{Is modularity crucial?} \label{sec:modular}
To evaluate the importance of architectural modularity, we design a monolithic variant of our model that eliminates the expert-based structure. In this baseline, the contextual representations from each modality are directly passed into the fusion module, bypassing the separation between modality-specific and multimodal experts. This modification leads to a notable degradation in performance, with the weighted F1-score dropping from $70.9\%$ to $67.8\%$ on the IEMOCAP dataset and from $69.5\%$ to $67.9\%$ on MELD. These results underscore a key insight: simply fusing contextual representations in a unified stream is insufficient. Instead, explicitly decoupling context modeling from cross-modal fusion, as implemented in \method{}, enables each component to specialize and contribute more effectively to the final prediction.

\begin{table*}[t!]
\centering
\resizebox{\textwidth}{!}{%
\begin{tabular}{@{}l|l|c|c|c| c@{}}
\toprule
\textbf{Method} & \textbf{\# Params} & \textbf{Text Feats.} & \textbf{Speech Feats.} & \textbf{IEMOCAP} & \textbf{MELD} \\
\midrule
MultiEmo~\citep{shi2023multiemo} & $\approx450$M & RoBERTa & OpenSMILE & $66.9\%$ & $65.3\%$ \\
~~+~~LLM/SLLM  &$\approx14$B & LLaMA & SALMONN & $69.5\%$ & $68.2\%$ \\ \midrule
HCAM~\citep{dutta2023hcam} & $\approx750$M & RoBERTa & wav2vec & $70.5\%$ & $65.8\%$\\
~~+~~LLM/SLLM &$\approx14$B & LLaMA & SALMONN & $70.6\%$ & $\underline{68.4}\%$\\ \midrule
MMGAT-EMO~\citep{zhang2025moe} & $\approx450$M & EmoBERTa & OpenSMILE & $65.5\%$ & $66.1\%$ \\ 
~~+~~LLM/SLLM &$\approx14$B & LLaMA & SALMONN & $68.6\%$ & $67.7\%$\\ \midrule \midrule
\textcolor{black}{\method{}~~w/o~~LLM/SLLM } &\textcolor{black}{$\approx450$M}& \textcolor{black}{RoBERTa} & \textcolor{black}{OpenSMILE} & \textcolor{black}{$69.5\%$} & \textcolor{black}{$66.4\%$} \\
\textcolor{black}{\method{}~~w/o~~LLM/SLLM }&\textcolor{black}{$\approx750$M}& \textcolor{black}{RoBERTa} & \textcolor{black}{wav2vec} & \textcolor{black}{$\mathbf{71.1}\%$} & \textcolor{black}{$\underline{68.4\%}$}\\
\method{} &$\approx14$B & LLaMA & SALMONN & $\underline{70.9\%}$ & $\mathbf{69.5\%}$\\
\bottomrule
\end{tabular}}
\caption{Comparison of some of the baseline methods \textcolor{black}{when re-designed with and without LLM features. The results show that the proposal is applicable even for non-LLM representations like OPENSMILE and wav2vec. } }
\label{tab:our_feats}
\end{table*}
\subsection{Are LLM  embeddings the panacea?}
An attentive reader might enquire whether the gains of our approach stem solely from the use of LLM/SLLM features. To isolate the impact of our architectural design from that of the input representations, we conduct controlled experiments where we retrain several prior ERC models using the same LLM and SLLM features employed in our proposed method. Since these baselines were originally optimized for different feature types, we perform our own hyperparameter tuning for each model to ensure a fair comparison (see Table~\ref{tab:our_feats}). Our findings show that while all baselines benefit to some extent from the improved representations, the performance of \method{} improves over previously published baselines. This highlights a key finding: the gains achieved by our framework are not solely attributable to powerful embeddings, but instead emerge from the modular architectural choices—particularly the separation of context modeling and cross-modal fusion, and the use of expert-level supervision and gating.\\
\textcolor{black}{To further assess whether the gains of \method{} are attributable to architectural design rather than encoder scale, we construct two variants of \method{} by replacing the LLM/SLLM encoders with more conventional pretrained representations: RoBERTa for text and \{OpenSMILE, wav2vec\} features for speech. Despite the substantial reduction in model capacity, both variants of \method{} consistently outperform all baseline methods that use the same feature representations (Table~\ref{tab:our_feats}). Notably, on IEMOCAP, \method{} with RoBERTa and wav2vec features achieves performance comparable to—and in some cases higher than—the corresponding LLM/SLLM-based configuration, suggesting improved generalization under reduced model capacity. These experiments demonstrate that the advantages of our approach are not tied to any specific set of pretrained embeddings, but rather stem from the modeling design that is proposed in this work.}



\subsection{Is decision-level MoE gating crucial?}\label{sec:gate_ablate}
To assess the importance of our decision-level Mixture-of-Experts (MoE) gating strategy, we conduct ablation studies by comparing \method{} against two architectural variants. In the first variant, termed \texttt{feat-MoE}, we shift the fusion point to the feature space—that is, the modality-specific and multimodal representations are combined before classification, rather than fusing their predictions. This design tests whether earlier integration is sufficient. In the second variant, called \texttt{No-Loss-MoE}, we eliminate the expert-level supervision altogether. Here, the model is trained solely on the final gated prediction using focal loss, omitting the individual loss signals for the audio, text, and multimodal experts.
As shown in Fig.~\ref{fig:moe}, both variants degrade performance: \texttt{feat-MoE} results in a $2.6\%$ drop on IEMOCAP and $0.6\%$ on MELD, while \texttt{No-Loss-MoE} leads to similar degradation. These findings reinforce two key design principles behind \method{}: (1) decision-level fusion is more effective than early feature fusion, likely because it preserves modality-specific discriminative cues; and (2) expert-specific supervision is essential for encouraging each expert to specialize in its respective modality or fusion pattern.
\subsection{Performance of Other LLMs/SLLMs} \label{sec:others}
We experiment with other speech and language large models (SLLMs and LLMs) for encoding speech and text, respectively. All models are trained using the same hyperparameters as \texttt{LLaMA-3.1-8B} and \texttt{SALMONN - 7B}. For the text modality, we experiment with \texttt{Qwen2.5-7B} and \texttt{Gemma-7B}, while for speech, we use \texttt{Qwen2-Audio-7B} and \texttt{SALMONN-13B}. 

\begin{figure*}[t!]
    \centering
    \begin{subfigure}[t]{0.48\textwidth}
        \centering
        \includegraphics[width=\linewidth,trim={1cm 6cm 13cm 2cm},clip]{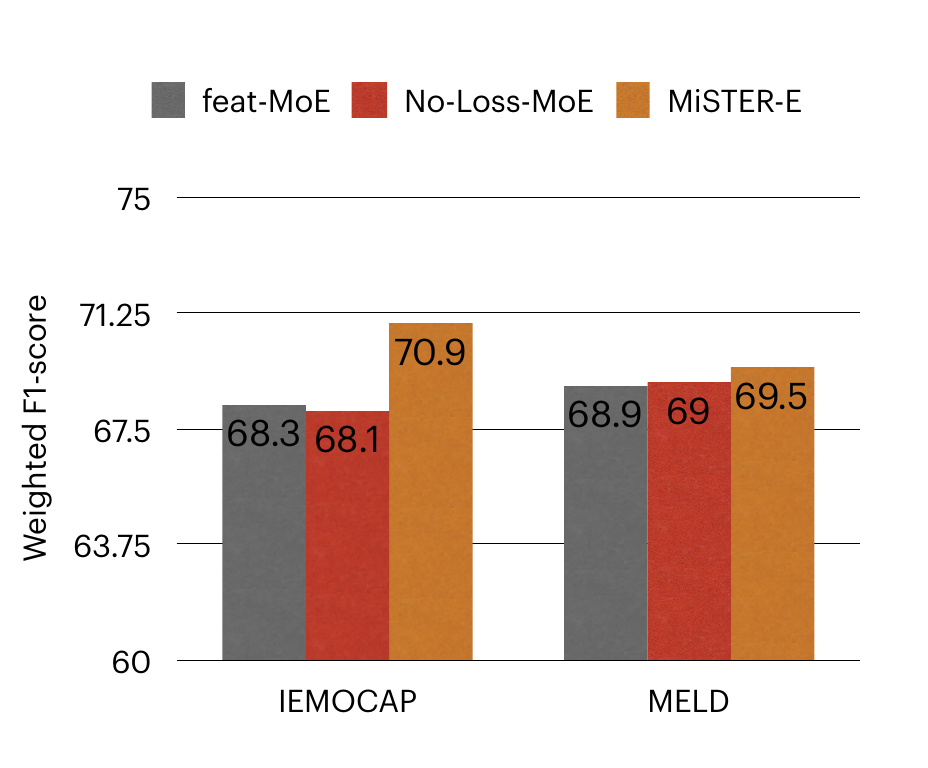}
        \caption{ Performance of \method{} with changes in the MoE gating strategy for IEMOCAP and MELD and }
        \label{fig:moe}
    \end{subfigure}\hfill
    \begin{subfigure}[t]{0.5\textwidth}
        \centering
        \raisebox{1.35\height}{%
        \resizebox{\linewidth}{!}{%
        \begin{tabular}{@{}l|l|c|c@{}}
\toprule
\textbf{Modality} & \textbf{Model} & \textbf{IEMOCAP} & \textbf{MELD} \\
\midrule
\multirow{3}{*}{\texttt{Text}} 
& \texttt{Gemma-7B} & $48.2\%$ & $58.1\%$ \\
& \texttt{Qwen2.5-7B} & $53.1\%$ & $65.5\%$ \\
& \texttt{LLaMA-3.1-8B} & $\mathbf{55.3\%}$ & $\mathbf{67.1\%}$ \\
\midrule \midrule
\multirow{3}{*}{\texttt{Speech}} 
& \texttt{SALMONN-7B} & $\mathbf{59.7\%}$  & $54.3\%$ \\
& \texttt{Qwen2-Audio-7B} & $59.2\%$ & $\mathbf{54.9\%}$ \\
& \texttt{SALMONN-13B} & $58.7\%$ & $53.2\%$ \\
\bottomrule
\end{tabular}}}
        \caption{Performance of different SLLMs and LLMs on the two datasets.}
        \label{tab:others}
    \end{subfigure}
    \caption{Comparison of model performance on IEMOCAP and MELD. 
(a) Effect of different MoE gating strategies in \method{}. 
(b) Performance of unimodal SLLMs and LLMs across the two datasets.}
\end{figure*}

As shown in Table~\ref{tab:others}, the performance of the speech encoders is comparable across models, with \texttt{SALMONN-7B} achieving the best result on IEMOCAP, and \texttt{Qwen2-Audio-7B} performing best on MELD. The larger \texttt{SALMONN-13B} underperforms compared to its smaller counterparts. For text, \texttt{LLaMA-3.1-8B} outperforms both \texttt{Gemma-7B} and \texttt{Qwen2.5-7B} on both datasets.

\section{Discussion}
\begin{figure*}[t!]
    \centering
\includegraphics[width=0.9\textwidth,trim={1cm 7cm 1cm 4cm},clip]{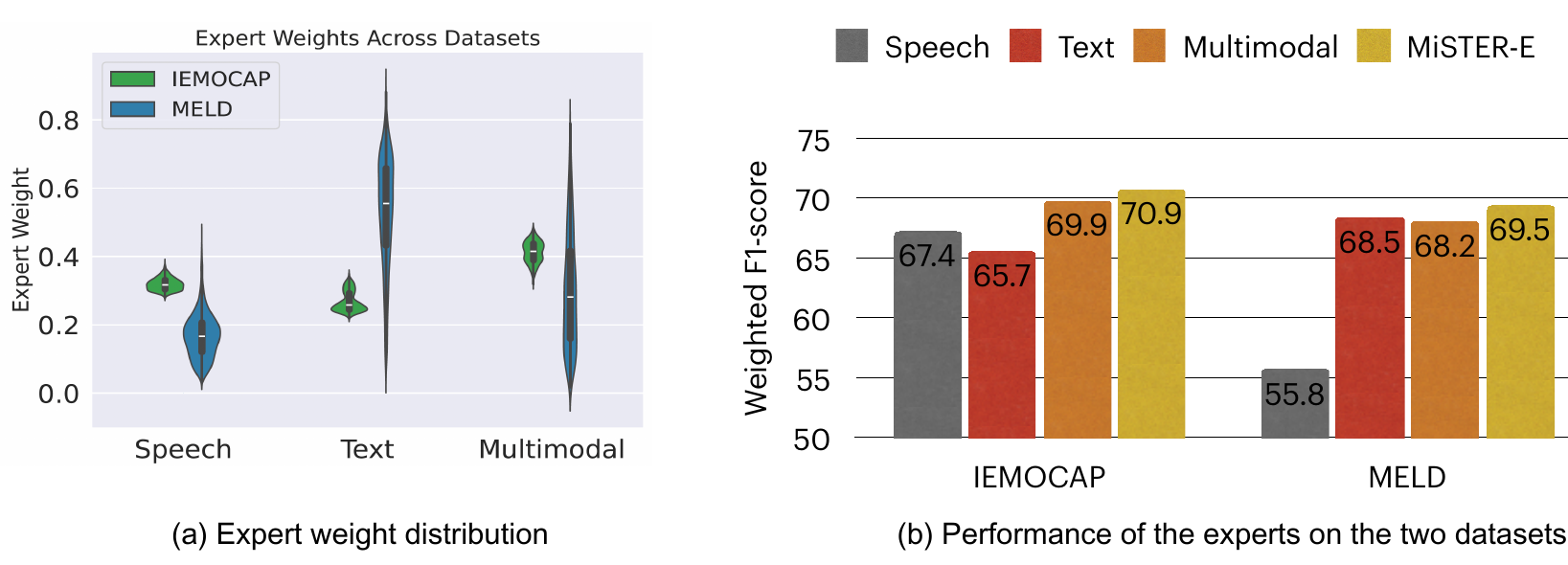}

    \caption{(a) Distribution of weights for the experts for the different datasets and (b) The performance of the experts and \method{} for the two datasets}
    \label{fig:comb_2}
\end{figure*}

\textcolor{black}{
\subsection{Expert Behavior and Modality Imbalance}}\label{sec:comb}
\textcolor{black}{
We analyze the behavior of the mixture-of-experts (MoE) gating mechanism on IEMOCAP and MELD to understand how the model adapts to dataset-specific modality characteristics. As shown in Fig.~\ref{fig:comb_2}(a), the gating network exhibits clear dataset-dependent preferences: it assigns higher weight to the multimodal expert for IEMOCAP, while favoring the text expert for MELD. This behavior reflects the relative reliability of modalities in the two datasets.\\
These trends are consistent with the performance of individual experts in Fig.~\ref{fig:comb_2}(b). On MELD, the speech expert performs substantially worse than the text expert, and the multimodal expert slightly underperforms the text expert due to the difficulty of aligning modalities with highly unequal predictive power. In contrast, on IEMOCAP, where speech and text modalities exhibit comparable performance, the multimodal expert effectively fuses both sources and outperforms the best unimodal expert by $2.5\%$. Across both datasets, \method{} consistently outperforms the strongest individual expert by approximately $1\%$, demonstrating the benefit of adaptive, per-utterance expert weighting.\\
The effect of expert consistency regularization aligns with these observations. For IEMOCAP, a moderate consistency strength ($\alpha=0.1$) improves performance from $70.4\%$ to $70.9\%$, while stronger regularization degrades performance. For MELD, where modality imbalance is pronounced, a small value ($\alpha=10^{-3}$) yields a modest gain of $0.1\%$, whereas larger values significantly reduce performance.  
}

\begin{figure*}[ht!]
    \centering
\includegraphics[width=\textwidth,trim={0cm 0cm 0cm 0cm},clip]{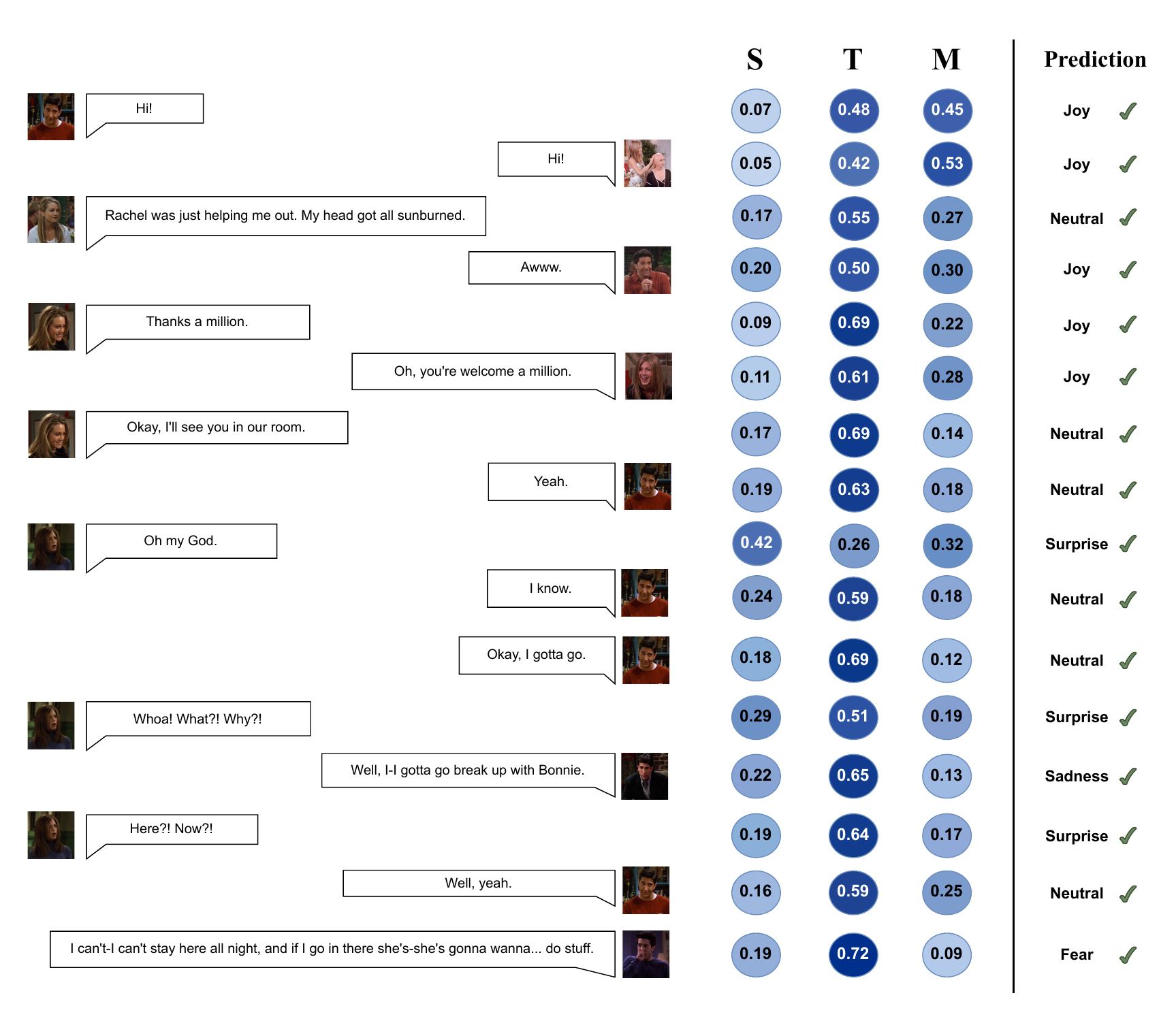}

    \caption{A case study from MELD where there are $16$ utterances in the conversation. The weights assigned to the different experts are shown. In most cases, text is assigned the highest weight (except utterance $2$ and $9$). S: Speech Expert, T: Text Expert, M: Multimodal Expert. The predictions of \method{} are also shown. The model correctly predicts all the utterances in the conversation, inspite of the emotion transitions.}
    \label{fig:case2}

\end{figure*}

\subsection{Class-wise Performance of \method{}}\label{sec:class_wise}
\begin{figure*}[t!]
    \centering
    \begin{subfigure}[b]{0.48\textwidth}
        \centering
        \includegraphics[width=\linewidth,trim={0cm 1cm 0cm 0cm},clip]{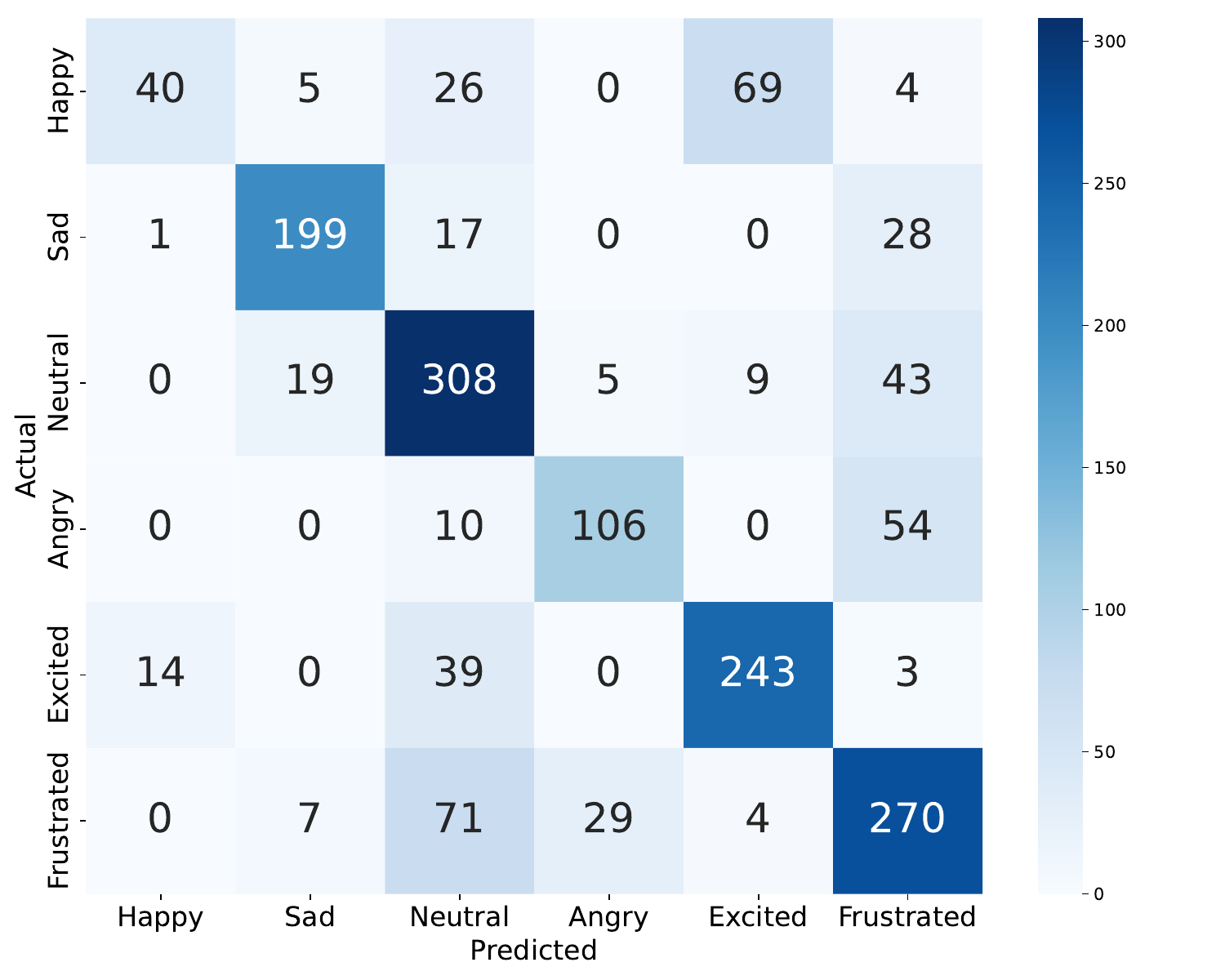}
        \caption{Confusion matrix of our model on the IEMOCAP dataset.}
        \label{fig:heatmap}
    \end{subfigure}\hfill
    \begin{subfigure}[b]{0.45\textwidth}
        \centering
        \resizebox{\linewidth}{!}{%
        \begin{tabular}{@{}l|c|l|c@{}}
        \toprule
        \multicolumn{2}{c|}{\textbf{IEMOCAP}} & \multicolumn{2}{c}{\textbf{MELD}} \\
        \midrule
        \textbf{Category} & \textbf{F1} & \textbf{Category}  & \textbf{F1}\\
        \midrule
        Angry & $68.4\%$ & Angry & $62.5\%$\\
        Excited & $77.9\%$ & Disgust & $42.9\%$ \\
        Frustrated & $69.0\%$ & Fear &  $30.6\%$\\
        Happy & $40.2\%$ & Joy & $65.4\%$ \\
        Neutral & $80.2\%$ & Neutral & $81.3\%$\\
        Sad & $83.8\%$ & Sad & $50.3\%$\\
        - & - & Surprise & $61.5\%$\\
        \bottomrule
        \end{tabular}}
        \caption{Per-class F1-scores for IEMOCAP and MELD.}
        \label{tab:class}
    \end{subfigure}
    \caption{(a) Confusion matrix for IEMOCAP. (b) Class-wise F1-scores on IEMOCAP and MELD, showing variation across different emotions.}
    \label{fig:heatmap_table}
\end{figure*}

The class-wise performance of our proposed method for the two datasets is shown in Table~\ref{tab:class}. We note that for IEMOCAP the worst performing class is ``happy'' - likely because it is the least frequent class and it is most confused with the ``excited'' category (Fig.~\ref{fig:heatmap}). Similarly, for MELD, the model performs most poorly on the ``fear'' category - again the least frequent class. However, we note that for both datasets, the model performs quite well on most emotion classes, thereby leading to an improved overall performance.

\subsection{Text-only Experiment}\label{sec:emory}

Although \method{} is designed for speech-text ERC, we evaluate its text-only variant to compare against recent LLM-based ERC approaches that frame the task as text generation and often rely on speaker metadata. We consider InstructERC~\cite{lei2023instructerc} as a representative baseline and adapt it to operate without speaker information for a fair comparison. Experiments are conducted on the EmoryNLP dataset~\cite{zahiri2018emotion}, which contains only textual transcripts across seven emotion classes.

\begin{figure*}[t!]
    \centering
\includegraphics[width=0.9\textwidth,trim={0cm 5cm 0cm 0cm},clip]{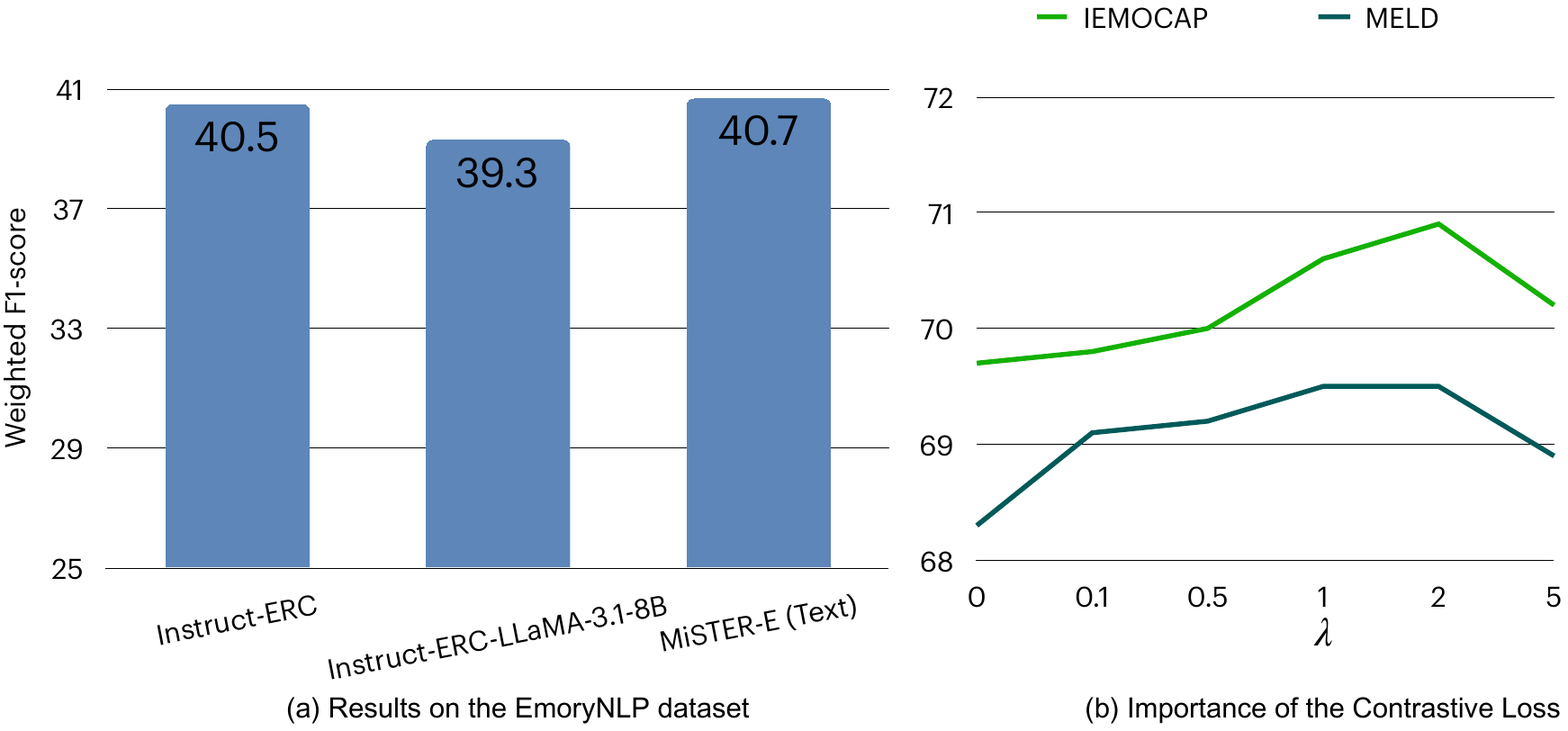}

    \caption{(a) Results of the text only experiment and, (b) Performance of \method{} with varying values of $\lambda$ (Eq.~\ref{eq:all}) for the two datasets. The validation set performance mirrors the trends of the test set. For IEMOCAP, validation accuracy is $63\%$ for $\lambda=0$ and $63.6\%$ for $\lambda=2$. For MELD, $\lambda=1$ yields the highest validation performance ($65.6\%$) compared to $65.4\%$ for $\lambda=0$}
    \label{fig:comb_3}
\end{figure*}

For this setting, text embeddings are extracted as described in Sec.~\ref{sec:unifeat}, followed by the Context Addition Network, while the speech branch and MoE gating losses are removed. As shown in Fig.~\ref{fig:comb_3}(a), \method{} achieves a modest improvement of $0.2\%$ over InstructERC. To control for model capacity, we further replace the \texttt{LLaMA-2-7B-chat} encoder used in InstructERC with \texttt{LLaMA-3.1-8B}. This variant performs worse than both InstructERC and \method{}, indicating that the observed gains stem from the proposed training and contextual modeling strategy rather than from the choice of LLM alone.



\subsection{Importance of the Contrastive Loss}

Figure~\ref{fig:comb_3}(b) analyzes the effect of the supervised contrastive loss on model performance. On MELD, incorporating the contrastive objective improves performance from $68.2\%$ ($\lambda=0$) to $69.5\%$ at $\lambda=1$, while on IEMOCAP performance increases from $70.2\%$ to $70.9\%$ at $\lambda=2$. These gains indicate that contrastive learning helps align emotion-relevant speech and text representations beyond what is achieved by focal loss alone. However, further increasing $\lambda$ leads to degraded performance on both datasets, suggesting that the contrastive loss serves as a beneficial auxiliary signal rather than a standalone supervisory objective.

\begin{figure*}[ht!]
    \centering
\includegraphics[width=\textwidth,trim={0cm 0cm 0cm 0cm},clip]{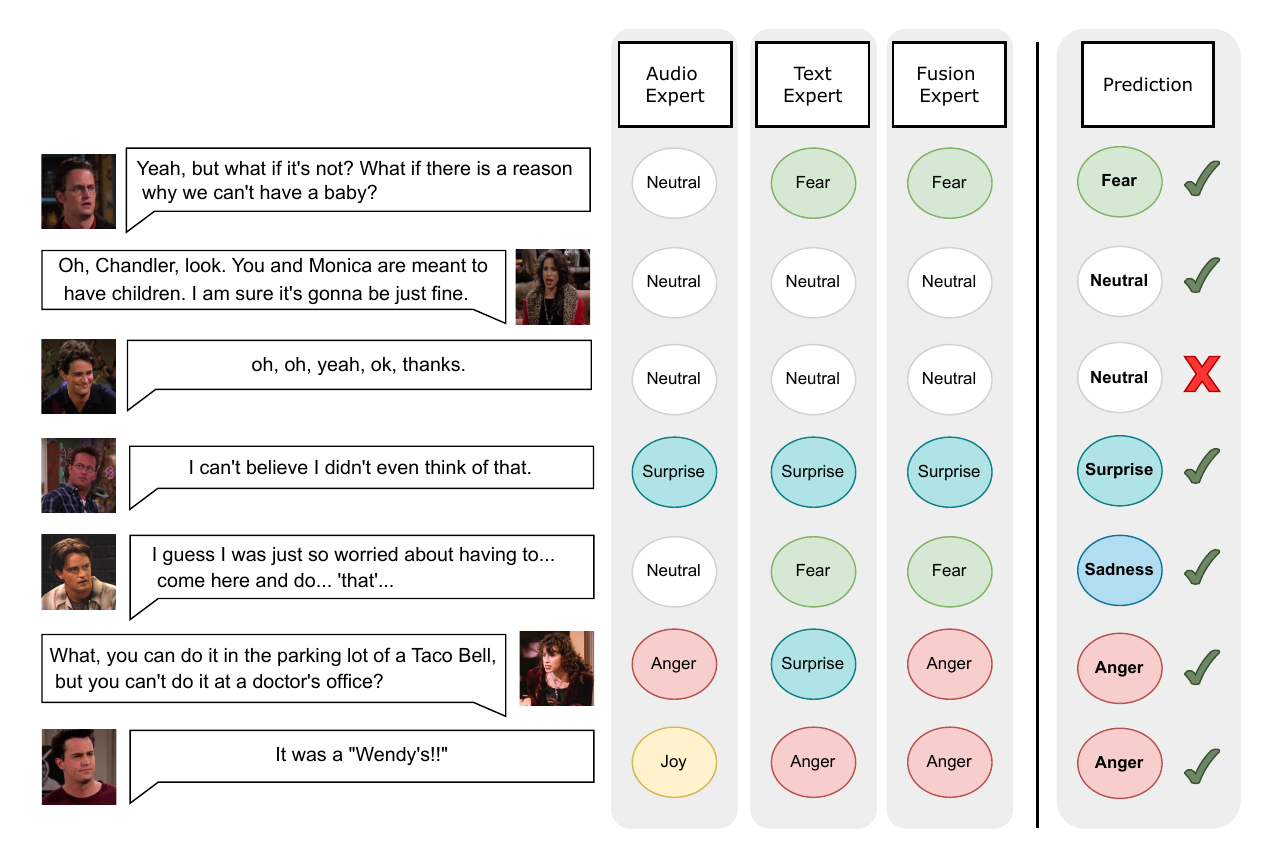}

    \caption{A case study from MELD where there are emotion shifts throughout the conversation. Out of the $7$ utterances in the conversation, \method{} correctly predicts $6$.}
    \label{fig:case1}

\end{figure*}
\subsection{Case Study}\label{sec:case}
We provide a case study from the MELD dataset in Fig.~\ref{fig:case1}. A number of key points that we wish to highlight from this example are as follows:
\begin{itemize}
    \item \method{} does not fail when there are emotion shifts. E.g., the conversation has $6$ emotions in $7$ utterances. Although, the model incorrectly predicts the third utterance to be neutral (instead of joy), it is able to predict the emotion shift from surprise to sadness to anger.
    \item Another interesting point is the output of the fifth utterance. None of the experts predict sadness class, yet the MoE strategy correctly marks the utterance as sad. This indicates the utility of the MoE gating strategy in our proposed method and differentiates it from static ensembling techniques.
\end{itemize}

\section{Conclusion}
We proposed \method{}, a modular framework for ERC that explicitly separates contextual modeling from multimodal fusion. Leveraging LLM-based representations for both speech and text, we model context in the conversations using a temporal inception block followed by a Bi-GRU, and perform modality fusion via an attention-based network. A Mixture-of-Experts (MoE) gate adaptively integrates decisions from context-aware and multimodal experts.
\method{} achieves new state-of-the-art performance on IEMOCAP, MELD, and CMU-MOSI without using speaker identity, demonstrating the efficacy of modular context-fusion and decision-level gating. \textcolor{black}{The auxiliary components such as focal loss, supervised contrastive learning, and KL-based consistency regularization provide modest performance improvements, and hence, they are best viewed as refinements that enhance training stability and robustness, rather than being essential elements of the proposed architecture.}

\textcolor{black}{\section{Limitations}}
\textcolor{black}{
While \method{} achieves strong performance across multiple benchmarks, several limitations remain,  
(i) the use of large LLM/SLLM encoders introduces non-trivial computational and memory overhead, which may limit applicability in low-resource or real-time settings, despite the use of parameter-efficient fine-tuning. We have partially addressed this concern in Table~\ref{tab:our_feats}, where the proposed approach is shown to be beneficial even for non-LLM based features,  
(ii) our evaluation is primarily conducted on benchmark datasets consisting of scripted or semi-scripted dialogues (e.g., TV shows and acted conversations), and performance under domain shift to spontaneous, real-world conversational settings remains an open question, (iii) emotion inference models are known to be susceptible to dataset biases related to demographic factors, language use, and annotation subjectivity and we have not performed any bias/fairness analysis in this study, and, (iv) throughout this work,  we intentionally avoid explicit speaker identity modeling, however, the utilization of speaker metadata may improve the modeling of interpersonal dynamics. Addressing some of these limitations  forms part of our future work. }

\section*{Declaration of generative AI and AI-assisted technologies in the writing process.}\noindent
During the preparation of this work the author(s) used ChatGPT in order to polish the writing. After using this tool/service, the author(s) reviewed and edited the content as needed and take(s) full responsibility for the content of the publication.

 \bibliographystyle{elsarticle-num} 
 \bibliography{references}







\end{document}